\documentclass[10pt,twocolumn,letterpaper]{article}

\usepackage[pagenumbers]{cvpr} % To force page numbers, e.g. for an arXiv version

\usepackage{amsmath,amsfonts,bm}

\def\eqref#1{equation~\ref{#1}}
\def\1{\bm{1}}

\def\vo{{\bm{o}}}

\def\vx{{\bm{x}}}
\def\vy{{\bm{y}}}

\def\evo{{o}}

\def\evx{{x}}
\def\evy{{y}}

\def\mS{{\bm{S}}}

\def\mX{{\bm{X}}}
\def\mY{{\bm{Y}}}

\DeclareMathAlphabet{\mathsfit}{\encodingdefault}{\sfdefault}{m}{sl}
\SetMathAlphabet{\mathsfit}{bold}{\encodingdefault}{\sfdefault}{bx}{n}

\usepackage{url}

\usepackage{graphicx}
\usepackage{epsfig}
\usepackage{amsmath}
\usepackage{amssymb}
\usepackage{booktabs}
\usepackage{algorithm}
\usepackage{algorithmicx}
\usepackage{algpseudocode}
\usepackage{multirow}
\usepackage{diagbox}
\usepackage{amsthm}
\usepackage{bm}
\usepackage{subcaption}

\usepackage[accsupp]{axessibility}
\usepackage{xcolor}
\definecolor{mycolor}{rgb}{0.01, 0.388, 0.288}
\usepackage[pagebackref,breaklinks,colorlinks]{hyperref}

\usepackage[capitalize]{cleveref}
\crefname{section}{Sec.}{Secs.}
\Crefname{section}{Section}{Sections}
\Crefname{table}{Table}{Tables}
\crefname{table}{Tab.}{Tabs.}

\def\cvprPaperID{1232} % *** Enter the CVPR Paper ID here
\def\confName{CVPR}
\def\confYear{2022}

\begin{document}
	\newcolumntype{L}[1]{>{\raggedright\arraybackslash}p{#1}}
	\newcolumntype{C}[1]{>{\centering\arraybackslash}p{#1}}
	\newcolumntype{R}[1]{>{\raggedleft\arraybackslash}p{#1}}
	\title{Exact Feature Distribution Matching for \\ Arbitrary Style Transfer and Domain Generalization}
	
\author{{Yabin Zhang$^{1}$, Minghan Li$^1$, Ruihuang Li$^1$, Kui Jia$^2$, Lei Zhang$^1$\thanks{Corresponding author} } \\
	$^1$Hong Kong Polytechnic University  \quad $^2$South China University of Technology \\
	{\tt\small \{csybzhang,csrhli,cslzhang\}@comp.polyu.edu.hk, liminghan0330@gmail.com, kuijia@scut.edu.cn}
}
%\author{{Yabin Zhang$^{1}$ \qquad Minghan Li$^1$ \qquad Ruihuang Li$^1$ \qquad Kui Jia$^2$ \qquad Lei Zhang$^1$\thanks{Corresponding author} } \\
%	$^1$Hong Kong Polytechnic University  \qquad $^2$South China University of Technology \\
%	{\tt\small \{csybzhang,csrhli,cslzhang\}@comp.polyu.edu.hk, liminghan0330@gmail.com, kuijia@scut.edu.cn}
%}

	\maketitle
	
	%%%%%%%%% ABSTRACT
	\begin{abstract}
		Arbitrary style transfer (AST) and domain generalization (DG) are important yet challenging visual learning tasks, which can be cast as a feature distribution matching problem. With the assumption of Gaussian feature distribution, conventional feature distribution matching methods usually match the mean and standard deviation of features. However, the feature distributions of real-world data are usually much more complicated than Gaussian, which cannot be accurately matched by using only the first-order and second-order statistics, while it is computationally prohibitive to use high-order statistics for distribution matching. 
		In this work, we, for the first time to our best knowledge, propose to perform Exact Feature Distribution Matching (EFDM) by exactly matching the empirical Cumulative Distribution Functions (eCDFs) of image features, which could be implemented by applying the Exact Histogram Matching (EHM) in the image feature space. Particularly, a fast EHM algorithm, named Sort-Matching, is employed to perform EFDM in a plug-and-play manner with minimal cost. The effectiveness of our proposed EFDM method is verified on a variety of AST and DG tasks, demonstrating new state-of-the-art results. Codes are available at \url{https://github.com/YBZh/EFDM}.
	\end{abstract}
	
	%%%%%%%%% BODY TEXT
	\section{Introduction} \label{Sec:introduction}
	
\begin{figure*}[h]
		\begin{center}
			\begin{subfigure}[b]{0.24\textwidth}
				\includegraphics[width=0.94\textwidth]{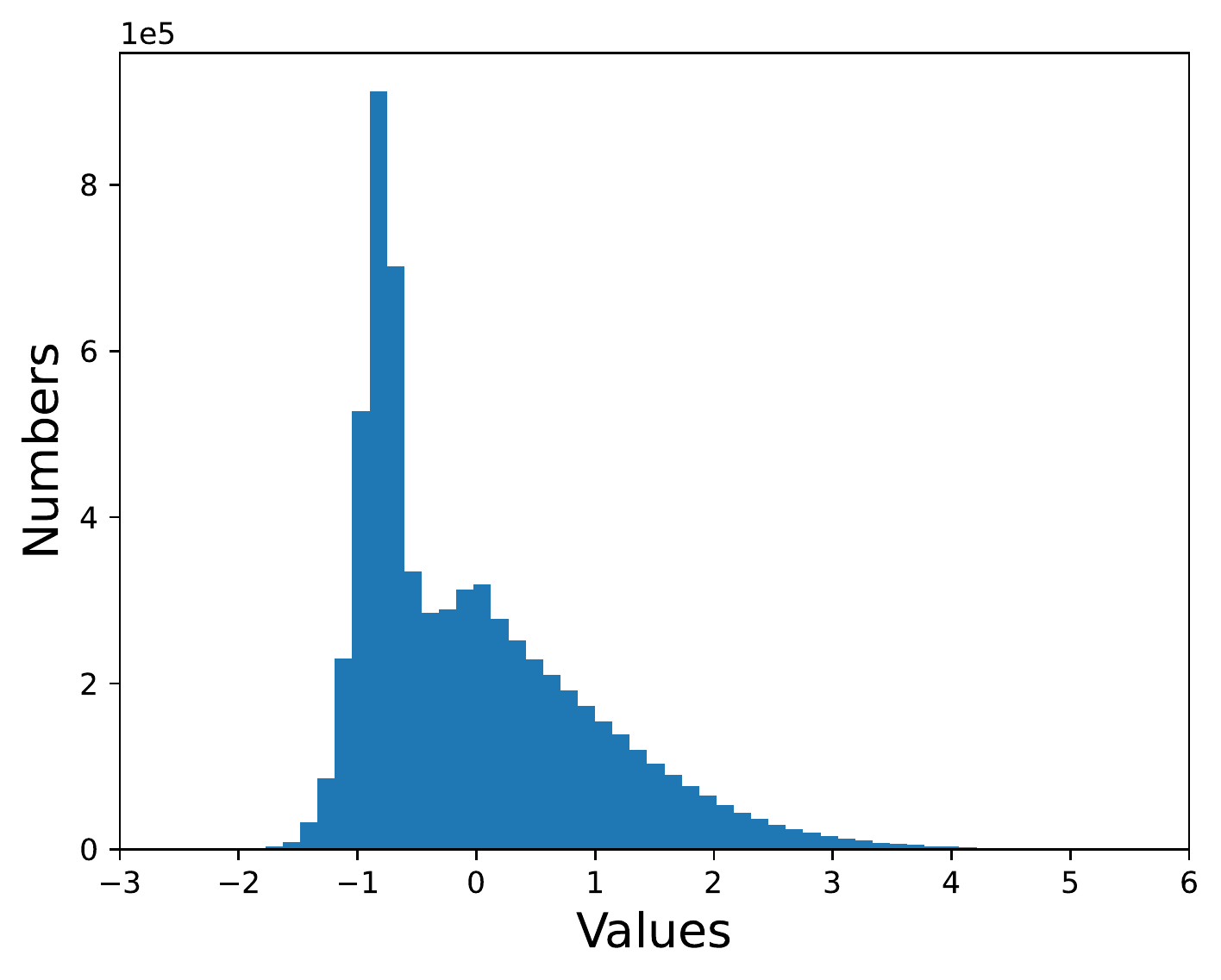}
				\caption{Art painting}
				\label{subFig:art_feat}
			\end{subfigure}%
			\begin{subfigure}[b]{0.24\textwidth}
				\includegraphics[width=0.94\textwidth]{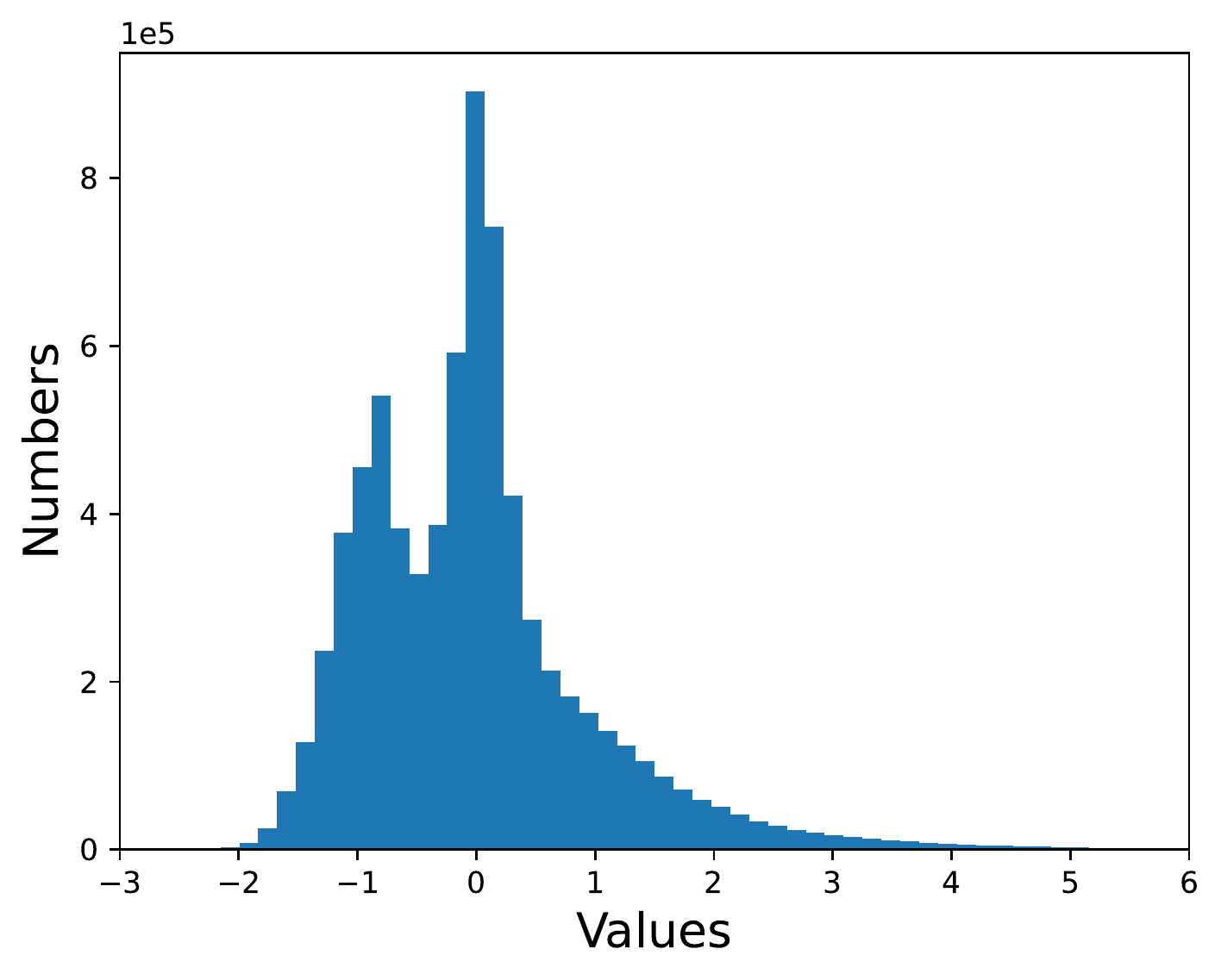}
				\caption{Cartoon}
				\label{subFig:cartoon_feat}
			\end{subfigure}%
			\begin{subfigure}[b]{0.24\textwidth}
				\includegraphics[width=0.94\textwidth]{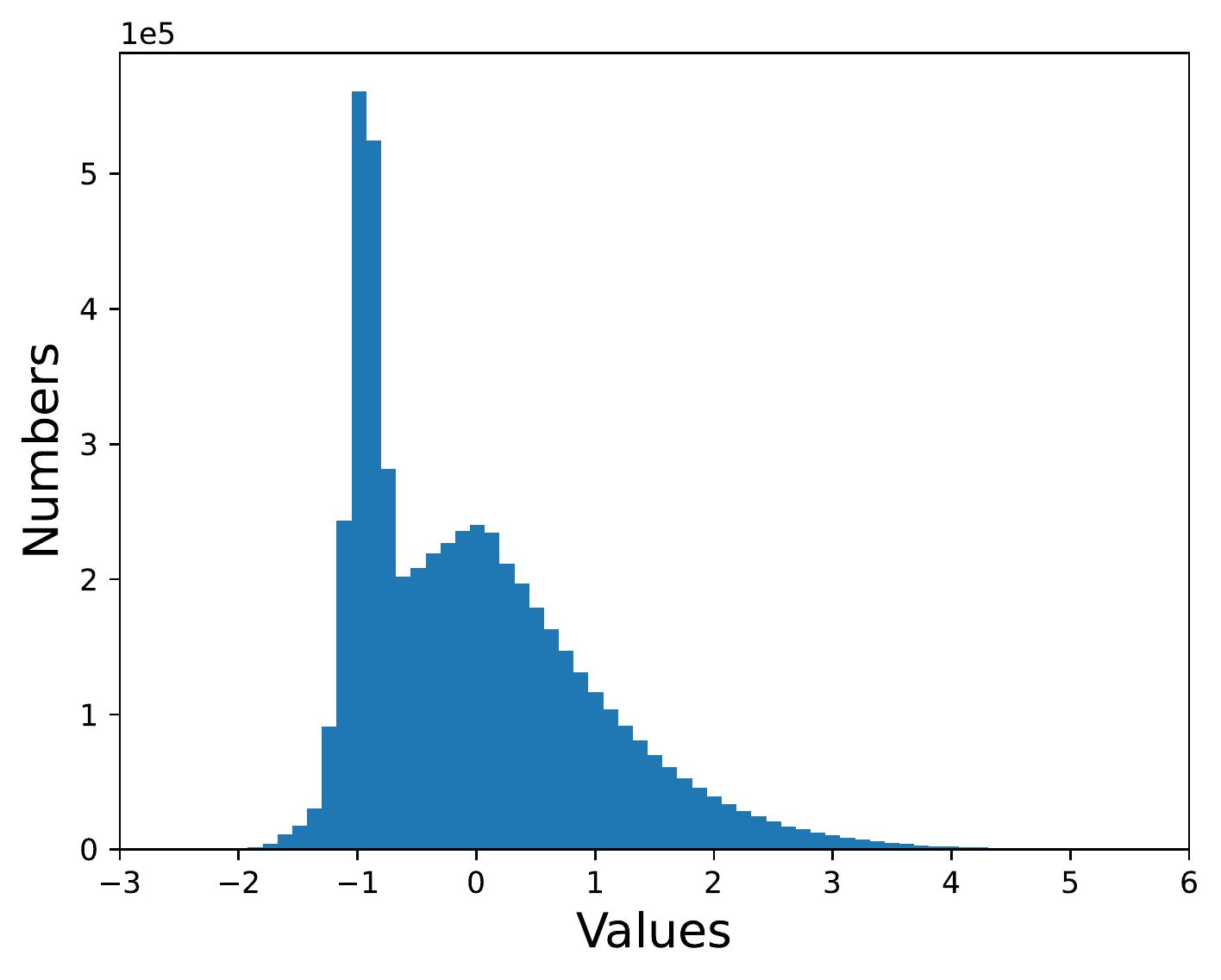}
				\caption{Photo} 	\label{subFig:photo_feat}
			\end{subfigure}%
			\begin{subfigure}[b]{0.24\textwidth}
				\includegraphics[width=0.96\textwidth]{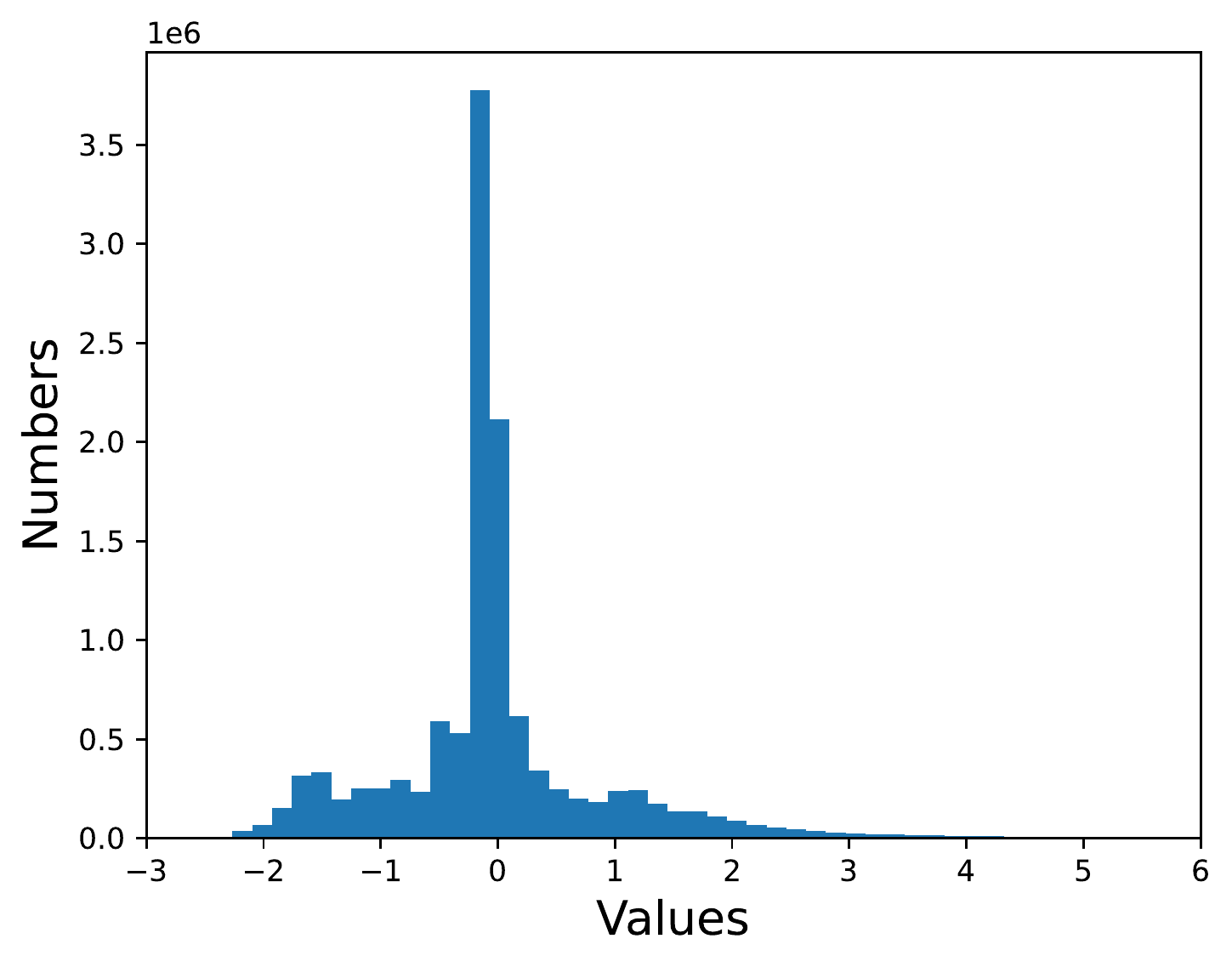}
				\caption{Sketch}	\label{subFig:sketch_feat}
			\end{subfigure}%
		\end{center}
		\vspace{-0.45cm}
		\caption{Histograms of feature values in a randomly selected channel, where features are computed from the first residual block of a ResNet-18 \cite{he2016deep} trained on the dataset of four domains \cite{li2017deeper}.  We first normalize the mean and standard deviation of each channel to be $0$ and $1$, respectively, and then collect feature values among all test samples in each domain for visualization. One can clearly see that the feature distributions of real-world data are usually too complicated to be modeled by Gaussian.
		} \label{Fig:feat_distribution_pacs}
		\vspace{-0.4cm}
	\end{figure*}

	Distribution matching is a long-standing statistical learning problem \cite{mood1950introduction}. With the popularity of deep models \cite{krizhevsky2012imagenet,he2016deep}, matching the distribution of deep features has attracted growing interest for its effectiveness in solving complex vision tasks. For instance, in arbitrary style transfer (AST) \cite{gatys2016image,huang2017arbitrary}, image styles can be interpreted as feature distributions and style transfer can be achieved by cross-distribution feature matching \cite{li2017demystifying,kalischek2021light}. Furthermore, by using style transfer techniques to augment training data, one can address the domain generalization (DG) tasks \cite{zhou2021domain,geirhos2018imagenet}, which target at generalizing the models learned in some source domains to other unseen domains.
	The most popular method of feature distribution matching is to match feature mean and standard deviation by assuming that features follow Gaussian distribution \cite{huang2017arbitrary,lu2019closed,mroueh2019wasserstein,li2019optimal,zhou2021domain}. Unfortunately, the feature distributions of real-world data are usually too complicated to be modeled by Gaussian, as illustrated in \cref{Fig:feat_distribution_pacs}. Therefore, feature distribution matching by using only mean and standard deviation is less accurate. It is desired to find more effective methods for more accurate and even Exact Feature Distribution Matching (EFDM).
	
\begin{figure*}[h]
		\begin{center}
			\includegraphics[width=0.85\textwidth]{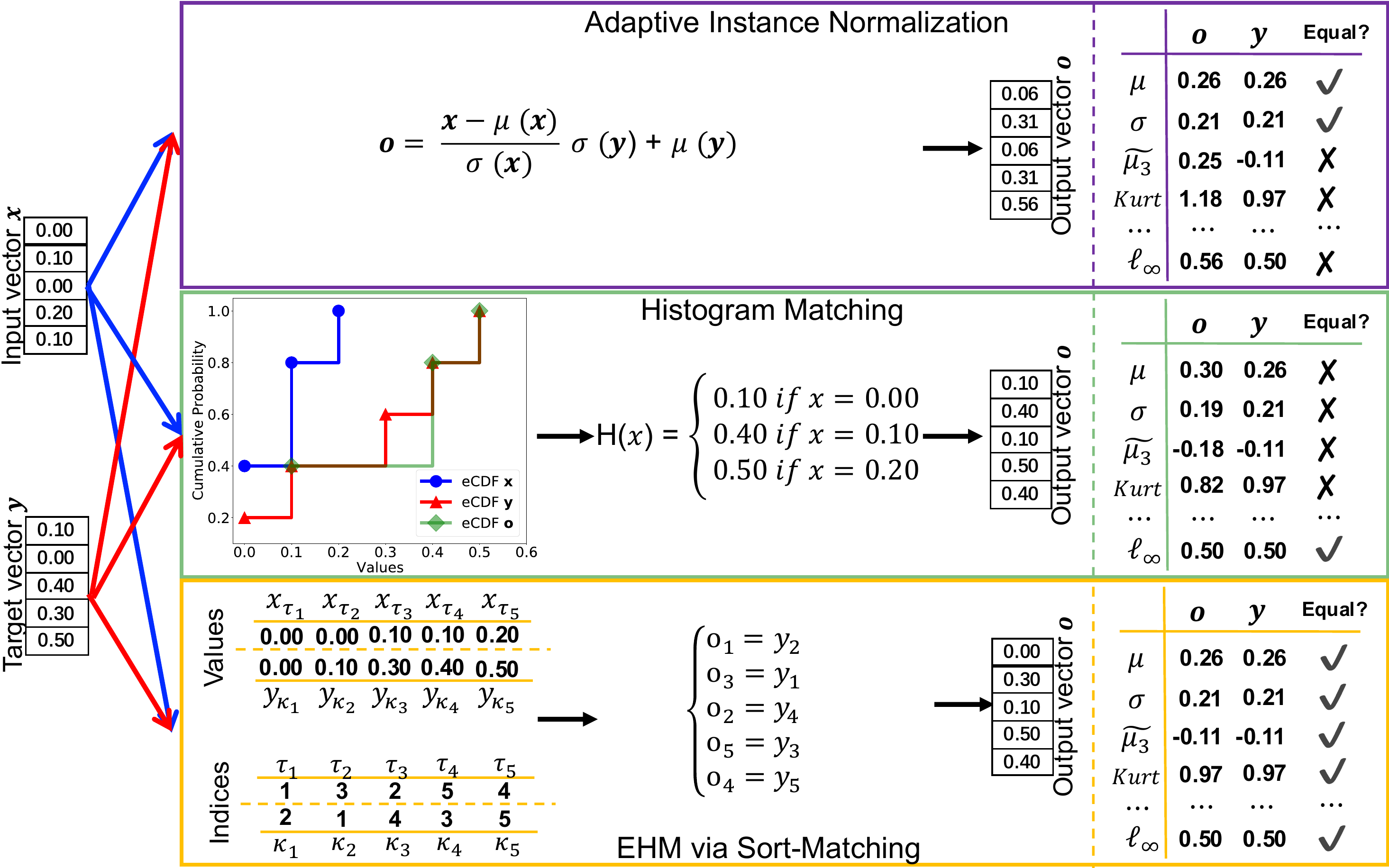}
			\vspace{-0.15cm}
			\caption{A comparison between AdaIN, HM and EHM via Sort-Matching using a toy example, where value precision is rounded to the level of 0.01. AdaIN only matches the mean and standard deviation between output vector $\vo$ and target vector $\vy$. Although the eCDF of $\vo$ is approximated to that of $\vy$ by HM, they are not exactly matched, leading to the mismatched distributions and, consequently, the mismatched statistics. The EHM via Sort-Matching exactly matches the eCDFs  of $\vo$ and $\vy$, resulting in exactly matched distributions and,  consequently, exactly matched statistics. 
				Notations of $\mu, \sigma, \widetilde{\mu_3}$,  $Kurt$ and $\ell_{\infty}$ indicate the mean, standard deviation, third standardized moment-skewness \cite{Skewness,joanes1998comparing}, fourth standardized moment-kurtosis \cite{Kurtosis,joanes1998comparing}, and infinite norm, respectively.
			} \label{Fig:comparison_HM_SM}
		\end{center}
		\vspace{-0.8cm}
	\end{figure*}

	Intuitively, EFDM can be done by matching the high-order statistics of features. Actually, high-order central moments have been explicitly introduced in \cite{kalischek2021light,zellinger2017central} to match distributions more precisely. However, considering high-order statistics in this way would introduce intensive computational overhead. Furthermore, the EFDM could only be theoretically achieved by matching central moments of infinite order \cite{zellinger2017central}, which is prohibitive in practice. Motivated by the Glivenko–Cantelli theorem \cite{van2000asymptotic}, which states that the empirical Cumulative Distribution Function (eCDF) asymptotically converges to the Cumulative Distribution Function when the  number  of  samples  approaches  infinity,
	%suggests that exact distribution matching could be achieved by matching empirical Cumulative Distribution Functions (eCDFs) exactly, 
	Risser \emph{et al.} \cite{risser2017stable} introduce the classical Histogram Matching (HM) \cite{histogram,gonzalez2002digital} method as an auxiliary measurement to minimize the feature distribution divergence. Unfortunately, HM can only approximately match eCDFs when there are equivalent feature values in inputs, since HM merges equivalent values as a single point and applies a point-wise transformation. (A toy example is illustrated in \cref{Fig:comparison_HM_SM}). This commonly happens for digital images with discrete integer values (\eg, 8-bits digital images). For features generated by deep models, equivalent feature values are also ineluctable due to their dependency on discrete image pixels and the use of activation functions, \eg, ReLU \cite{nair2010rectified} and ReLU6 \cite{krizhevsky2010convolutional} (please refer to \cref{Fig:percent_equivalent} for more details). All these facts impede the effectiveness of EFDM via HM.
	
	To solve the above mentioned problem, we, for the first time to our best knowledge, propose to perform EFDM by exactly matching the eCDFs of image features, resulting in exactly matched feature distributions (when the number of samples approaches infinity) and consequently exactly matched mean, standard deviation, and high-order statistics (see the toy example in \cref{Fig:comparison_HM_SM}). The exact matching of eCDFs can be implemented by applying the Exact Histogram Matching (EHM) algorithm \cite{hall1974almost,coltuc2006exact} in the feature space. Specifically, by distinguishing the equivalent feature values and applying an element-wise transformation, EHM conducts more fine-grained and more accurate matching of eCDFs than HM. In this paper, a fast EHM algorithm, named Sort-Matching \cite{rolland2000fast}, is adopted to perform EFDM in a plug-and-play manner with minimal cost. 
	
	%To solve the above mentioned problem, we, for the first time to our best knowledge, propose to perform EFDM by exactly matching the eCDFs of image features. Actually, by distinguishing the equivalent feature values and applying a value-wise transformation, we could do fine-grained and more accurate matching of eCDFs than HM. As proofed in the Glivenko–Cantelli theorem \cite{van2000asymptotic}, exactly matched eCDFs will result in exactly matched distributions when the number of samples approaches infinity, and consequently result in the exactly matched mean, standard deviation, and high-order statistics (see the toy example illustrated in \cref{Fig:comparison_HM_SM}). The exact matching of eCDFs can be implemented by applying the Exact Histogram Matching (EHM) algorithm \cite{hall1974almost,coltuc2006exact} in the image feature space. While EHM can be conducted with different strategies, we empirically find that they yield similar results in the AST and DG applications.	Particularly, a fast EHM algorithm, named Sort-Matching \cite{rolland2000fast}, is adopted in our work to perform EFDM in a plug-and-play manner with minimal cost. With EFDM, more faithfully style-transferred outputs can be obtained in AST, and more diverse data augmentations can be achieved in DG, resulting in new state-of-the-art results with high efficiency. 
	
	With EFDM, we conduct cross-distribution feature matching in one shot (cf. \cref{Equ:ast_EFDM}) and propose a new style loss  (cf. \cref{Equ:style_loss}) to more accurately measure distribution divergence, producing more stable style-transfer images in AST. Following \cite{zhou2021domain}, we extend EFDM to generate feature augmentations with mixed styles, leading to the Exact Feature Distribution Mixing (EFDMix) (cf. \cref{Equ:mixsorting}), which can provide more diverse feature augmentations for DG applications. Our method achieves new state-of-the-arts on a variety of AST and DG tasks with high efficiency.

	%The organization of this paper is as follows. We first review the methods of AST, DG, and EHM in \cref{Sec:related_work}. In \cref{Sec:methodology}, we perform EFDM by matching eCDFs of image features and apply EFDM to tasks of AST and DG. Finally, we justify the effectiveness of EFDM with empirical results on AST and DG, and conduct in-depth analyses in \cref{Sec:experiments}. 

	%\begin{theorem} 
	%	(Glivenko–Cantelli theorem). Assume that $x_1$, $x_2$, $\cdots$, $x_n$ are independent and identically-distributed random variables in $\mathbb{R}$ with the cumulative distribution function $F(x)$, we have 
	%	\begin{equation}
	%	\| \widehat{F} - F \| = \sup_{x\in \mathbb{R}} \left| \widehat{F}(x) - F(x)  \right| \rightarrow 0 
	%	\end{equation}
	%	when the sample number $n$ approaches infinity. $\widehat{F}(x)$ is the empirical cumulative distribution function:
	%	\begin{equation}
	%	\widehat{F}(x) = \frac{1}{n} \sum_{i=1}^{n} \mathbf{1}_{x_i \leq x},
	%	\end{equation}
	%	where $\mathbf{1}_{A}$ is the indicator of event $A$.   \label{theorem}
	%\end{theorem}

	\section{Related Work} \label{Sec:related_work}
	
	\noindent\textbf{Arbitrary style transfer} (AST) has been investigated in two conceptual directions: iterative optimization-based methods and feed-forward methods. The former \cite{gatys2016image,risser2017stable,kalischek2021light} optimize image pixels in an iterative manner, whereas the latter \cite{huang2017arbitrary,li2017universal,lu2019closed, mroueh2019wasserstein,li2019optimal} generate style-transferred output in one shot. 
	Our method belongs to the latter one, which is generally faster and suitable for real-time applications. In both directions, transferring styles can be interpreted as a problem of feature distribution matching by assuming the image styles can be represented by feature distributions. Specifically, the seminal work in \cite{gatys2016image} adopts the second-order moments captured by the Gram matrix as the style representation. The loss introduced in  \cite{gatys2016image} is rewritten as a Maximum Mean Discrepancy between image features in \cite{li2017demystifying}, bridging style transfer and feature distribution matching.  Actually, many AST methods can be interpreted from the perspective of feature distribution matching. Based on the Gaussian prior assumption, feature distribution matching is conducted by matching mean and standard deviation in AdaIN \cite{huang2017arbitrary}. Compared to AdaIN, WCT \cite{li2017universal} additionally considers the covariance of feature channels via a pair of feature transforms, whitening and coloring. By additionally taking the content loss in \cite{gatys2016image} into the framework of WCT, a closed-form solution is presented in \cite{lu2019closed, mroueh2019wasserstein,li2019optimal}. Besides the widely used first and second order feature statistics, high-order central moments and HM are introduced in \cite{kalischek2021light} and \cite{risser2017stable} respectively for more exact distribution matching  by relaxing the assumption of Gaussian feature distributions. However, computing high-order statistics explicitly introduces intensive computational overhead and the EFDM via HM is impeded by equivalent feature values. To this end, we, for the first time to our best knowledge, propose an accurate and efficient way for EFDM by exactly matching the eCDFs of image features, leading to more faithful AST results (please refer to \cref{Fig:view_style} for visual examples).

	\noindent\textbf{Domain generalization} (DG) aims to develop models that can generalize to unseen distributions. Typical DG methods include learning domain-invariant feature representations \cite{motiian2017unified,li2018domain,gong2019dlow,zhang2019domain,chen2019progressive,zhang2020unsupervised,zhao2020domain}, meta-learning-based learning strategies \cite{li2018learning,du2020learning,chen2022comen}, data augmentation \cite{volpi2018generalizing,yue2019domain,zhou2020learning,geirhos2018imagenet,nam2021reducing,zhou2021domain} and so on \cite{wang2021generalizing,zhou2021survey}. Among all above methods, the recent state-of-the-art \cite{zhou2021domain} augments cross-distribution features based on the feature distribution matching technique \cite{huang2017arbitrary}, which is introduced in the above AST part. By utilizing high-order statistics implicitly via the proposed EFDM method, more diverse feature augmentations can be achieved and significant performance improvements have been observed (please refer to  \cref{tab:321_pacs,Tab:reid} for details).

	\noindent\textbf{Exact histogram matching} (EHM) was proposed to match histograms of image pixels exactly. Compared to classical HM, EHM algorithms distinguish equivalent pixel values either randomly \cite{rosenfeld1976digital,rolland2000fast} or according to their local mean \cite{hall1974almost,coltuc2006exact}, leading to more accurate matching of histograms. The difference between outputs of EHM and HM in image pixel space is typically small, which is hardly perceptible to human eyes. However, this small difference can be amplified in the feature space of deep models, leading to clear divergence in feature distribution matching. We hence propose to perform EFDM by exactly matching the eCDFs of image features via EHM. While EHM can be conducted with different strategies, we empirically find that they yield similar results in our applications, and thus we promote the fast Sort-Matching \cite{rolland2000fast} algorithm for EHM. 
	%	\vspace{-0.2cm}
	
	%improves over HM by distinguishing equivalent values, resulting in the exact eCDF matching. Randomly distinguishing equivalent values \cite{rosenfeld1976digital} and making distinguishing according to their local mean \cite{hall1974almost,coltuc2006exact} are two popular methods. Actually, the divergence between outputs of EHM and HM is typically small. Such a small divergence is hardly perceptible to human eyes in image pixel space. However, this small divergence can be amplified with the deepening of deep models, leading to clear advantages of EHM over HM in feature distribution matching. Additionally, most EHM methods introduce additional computational overhead over HM, e.g., calculating the local mean \cite{hall1974almost,coltuc2006exact}, impeding their practical usage. In contrast, our promoted Sort-Matching \cite{rolland2000fast} is faster than HM and could be conducted with the complexity of $O(n\log n)$, facilitating its application in deep models.

	\section{Methodology} \label{Sec:methodology}
	\vspace{-0.05cm}
	%We first introduce the adaptive instance normalization and HM, as well as their deficiencies in practical usage. To tackle these issues, we introduce the EHM and present how to practically apply EHM in tasks of AST and DG.  
	
	\subsection{AdaIN, HM and EHM}
	\vspace{-0.05cm}
	\noindent\textbf{Adaptive instance normalization} (AdaIN) \cite{huang2017arbitrary} transforms an input vector $\vx \in \mathbb{R}^n$,  which is sampled from a random variable $X$, into an output vector $\vo \in \mathbb{R}^n$, whose mean and standard deviation match those of a target vector $\vy \in \mathbb{R}^m$  sampled from a random variable $Y$:
	\begin{align}
	\vo = \frac{\vx - \mu(\vx)}{\sigma(\vx)} \sigma(\vy)  + \mu(\vy),
	\end{align}
	where $\mu(\cdot)$ and $\sigma(\cdot)$ indicate the mean and standard deviation of referred data, respectively.
	By assuming that $X$ and $Y$ follow Gaussian distributions and $n$ and $m$ approach infinity, AdaIN can achieve EFDM by matching feature mean and standard deviation \cite{lu2019closed,mroueh2019wasserstein,li2019optimal}. 
	However, feature distributions of real-world data usually deviate much from Gaussian, as can be seen from \cref{Fig:feat_distribution_pacs}. Therefore, matching feature distributions by AdaIN is less accurate. 
	
	\noindent\textbf{Histogram matching} (HM) \cite{histogram,gonzalez2002digital} aims to transform an input vector $\vx$ into an output vector $\vo$, whose eCDF matches the target eCDF of a target vector $\vy$. The eCDFs of $\vx$ and $\vy$ are defined as:
	\begin{equation} \label{equ:cdf}
	\widehat{F}_{X}(x) = \frac{1}{n} \sum\nolimits_{i=1}^{n} \mathbf{1}_{x_i \leq x}, \quad \widehat{F}_{Y}(y) = \frac{1}{m} \sum\nolimits_{i=1}^{m} \mathbf{1}_{y_i \leq y},
	\end{equation}
	where $\mathbf{1}_{A}$ is the indicator of event $A$ and $x_i$ (or $y_i$) is the $i$-th element of $\vx$ (or $\vy$). 
	For each element $x_i$ of the input vector $\vx$, we find the $y_j$ that satisfies $\widehat{F}_{X}(x_i) = \widehat{F}_{Y}(y_j)$, resulting in the transformation function: $H(x_i)=y_j$.
	One may opt to match the explicit histograms as in discrete image space \cite{gonzalez2002digital}. It is worth mentioning that matching eCDFs is equivalent to matching histograms with bins of infinitesimal width, which is however hard to achieve due to the finite number of bits to represent features.
	
\begin{figure}[htp]
		\begin{center}
			\includegraphics[width=0.9\linewidth]{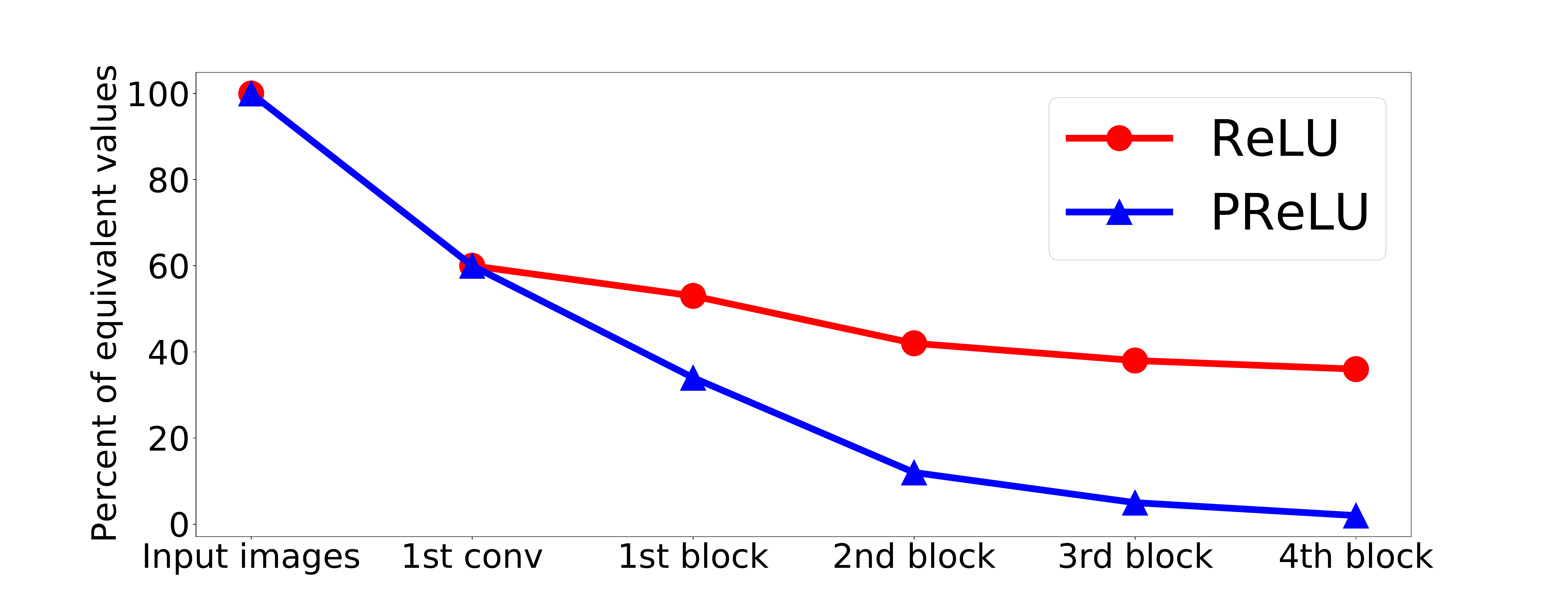}
			\caption{An illustration of the percent of equivalent feature values (\ie, $\frac{\textrm{number of equivalent  values}}{\textrm{number of all values}} * 100$) in ResNet18 feature maps with an input image of resolution $224\times 224$ . `1st conv' represents the output of the first convolution layer. `1st block', `2nd block', `3rd block', and `4th block' indicate the outputs of the 1st, 2nd, 3rd, and 4th residual blocks, respectively. `ReLU' and `PReLU' indicate the vanilla ResNet18 with ReLU \cite{nair2010rectified} and PReLU \cite{he2015delving} activation functions, respectively. The percentage depends on the number of bits to represent the feature values and the size of feature maps. In the original image pixel space, the percentage is close to 100 percent since the pixels are quantized into 8-bits. The percentage decreases in the floating number (32-bits) feature space, as well as the depth of blocks since deeper blocks have smaller feature maps. In addition,	compared to PReLU, there are generally more equivalent feature values for models with ReLU, since the ReLU function sets all negative values to zero.  
			} \label{Fig:percent_equivalent}
		\end{center}
		\vspace{-0.3cm}
	\end{figure}

	Ideally, HM could exactly match eCDFs of image features in the continuous case. 
	Unfortunately, HM can only approximately match eCDFs when there exist equivalent feature values in inputs, since HM merges equivalent values as a single point and applies a point-wise transformation (please refer to the toy example in \cref{Fig:comparison_HM_SM}). For features generated by deep models, equivalent feature values are common due to their dependency on discrete image pixels and the use of activation functions, \eg, ReLU \cite{nair2010rectified} and ReLU6 \cite{krizhevsky2010convolutional} (please refer to \cref{Fig:percent_equivalent} for more details). All these facts impede the effectiveness of EFDM via HM.

	\noindent\textbf{Exact Histogram Matching} (EHM) \cite{hall1974almost,coltuc2006exact} was proposed to match histograms of image pixels exactly. Different from HM, EHM algorithms distinguish equivalent pixel values and apply an element-wise transformation so that a more accurate histogram matching can be achieved. 
	While EHM can be conducted with different strategies, we adopt the Sort-Matching algorithm \cite{rolland2000fast} for its fast speed. 
	Sort-Matching is based on the quicksort strategy \cite{sedgewick1978implementing}, which is generally accepted as the fastest sort algorithm with complexity of $O(n\log n)$. 
	%proposed to tackle the histogram mismatching (i.e., eCDF mismatching) problem in the image pixel space. To match eCDFs exactly, EHM distinguishes  equivalent values via a strict sort strategy and could transform equivalent values to different outputs, as illustrated in \cref{Fig:comparison_HM_SM}. Different EHM methods are distinguished by their sort strategies of equivalent values. 
	%This paper employs the Sort-Matching method \cite{rolland2000fast} based on the quicksort strategy \cite{sedgewick1978implementing}, which is the generally accepted fastest sort algorithm with the complexity of $O(n\log n)$. 
	As stated by its name, Sort-Matching is implemented by matching two sorted vectors, whose indexes are illustrated in a one-line notation \cite{bogart1989introductory} as:
	\begin{equation} \label{equ:two_line_permu}
	\begin{split}
	&\vx: \tau = \left( \begin{array}{ccccc}
	\tau_1 & \tau_2 & \tau_3 & \cdots & \tau_n
	\end{array}  \right),  \\ 	&\vy: \kappa = \left( \begin{array}{ccccc}
	\kappa_1 & \kappa_2 & \kappa_3 & \cdots & \kappa_n
	\end{array}  \right),
	\end{split}
	\end{equation}
	where $\{ x_{\tau_i}\}_{i=1}^n$ and $\{y_{\kappa_i}\}_{i=1}^n$ are sorted values of $\vx$ and $\vy$ in ascending order.
	%where the top and bottom rows indicate the value indexes before sort and after sort, respectively. 
	In other words, $\evx_{\tau_1} = \min(\vx)$, $\evx_{\tau_n} = \max(\vx)$, and $\evx_{\tau_i} \leq \evx_{\tau_j}$ if $i<j$.  $\evy_{\kappa_i}$ is similarly defined.
	Based on the definition in \cref{equ:two_line_permu}, Sort-Matching outputs $\vo$ with its $\tau_i$-th element $\evo_{\tau_i}$ as:
	\begin{equation} \label{Equ:sorting_matching}
	\evo_{\tau_i} = \evy_{\kappa_i}.
	\end{equation}
	Compared to AdaIN, HM and other EHM algorithms \cite{hall1974almost,coltuc2006exact}, Sort-Matching additionally assumes that the two vectors to be matched are of the same size, \ie $m=n$, which is satisfied in our focused applications of AST and DG. In other applications where the two vectors are of different sizes, interpolation or dropping elements can be conducted to make $\vy$ and $\vx$ the same size.
	
	\subsection{EFDM for AST and DG}  \label{subsec:apply_sort_matching}
	
	In this section, we apply EFDM to tasks of AST and DG. We conduct the exact eCDFs matching by applying the EHM algorithm via Sort-Matching in the image feature space. 
	To enable the gradient back-propagation in deep models, we practically perform EFDM by modifying \cref{Equ:sorting_matching} as:
	\begin{equation}  \label{Equ:sorting_matching_enable_gradient}
	\textrm{EFDM}(\vx,\vy): \evo_{\tau_i} = \evx_{\tau_i} +  \evy_{\kappa_i}   - \langle \evx_{\tau_i} \rangle,
	\end{equation}
	where $\langle \cdot \rangle$ represents the stop-gradient operation \cite{chen2021exploring}.  We stop the gradients to the style feature $\evy_{\kappa_i}$ following \cite{huang2017arbitrary,zhou2021domain}.  
	Given the input data $\mX \in \mathbb{R}^{B\times C \times HW}$ and the style data $\mY \in \mathbb{R}^{B\times C \times H W}$, we apply EFDM in a channel-wise manner following \cite{huang2017arbitrary,zhou2021domain}, where $B,C,H,W$ indicate batch size, channel dimension, height, and width, respectively.
	The proposed EFDM does not introduce any parameters and can be used in a plug-and-play manner with few lines of codes and minimal cost, as summarized in \cref{alg:sortingmatching}. 
	
	\begin{algorithm}[htp]
		\caption{PyTroch-like pseudo-code for EFDM.} \label{alg:sortingmatching}
		\textcolor{mycolor}{\# $\vx$, $\vy$: input and target vectors  of the same shape (n)} \\
		\_, IndexX = torch.sort($\vx$) \qquad\quad \qquad \textcolor{mycolor}{\# Sort $\vx$ values} \\ 
		SortedY, \_ = torch.sort($\vy$)\qquad\quad \qquad \textcolor{mycolor}{\# Sort $\vy$ values} \\   
		InverseIndex = IndexX.argsort(-1)  \textcolor{mycolor} \\%{\#  Prepare inverse\_index of $\vx$}  \\
		return $\vx$ + SortedY.gather(-1, InverseIndex) - $\vx$.detach()
	\end{algorithm}

	\begin{figure}[htp]
	\begin{center}
		\includegraphics[width=0.99\linewidth]{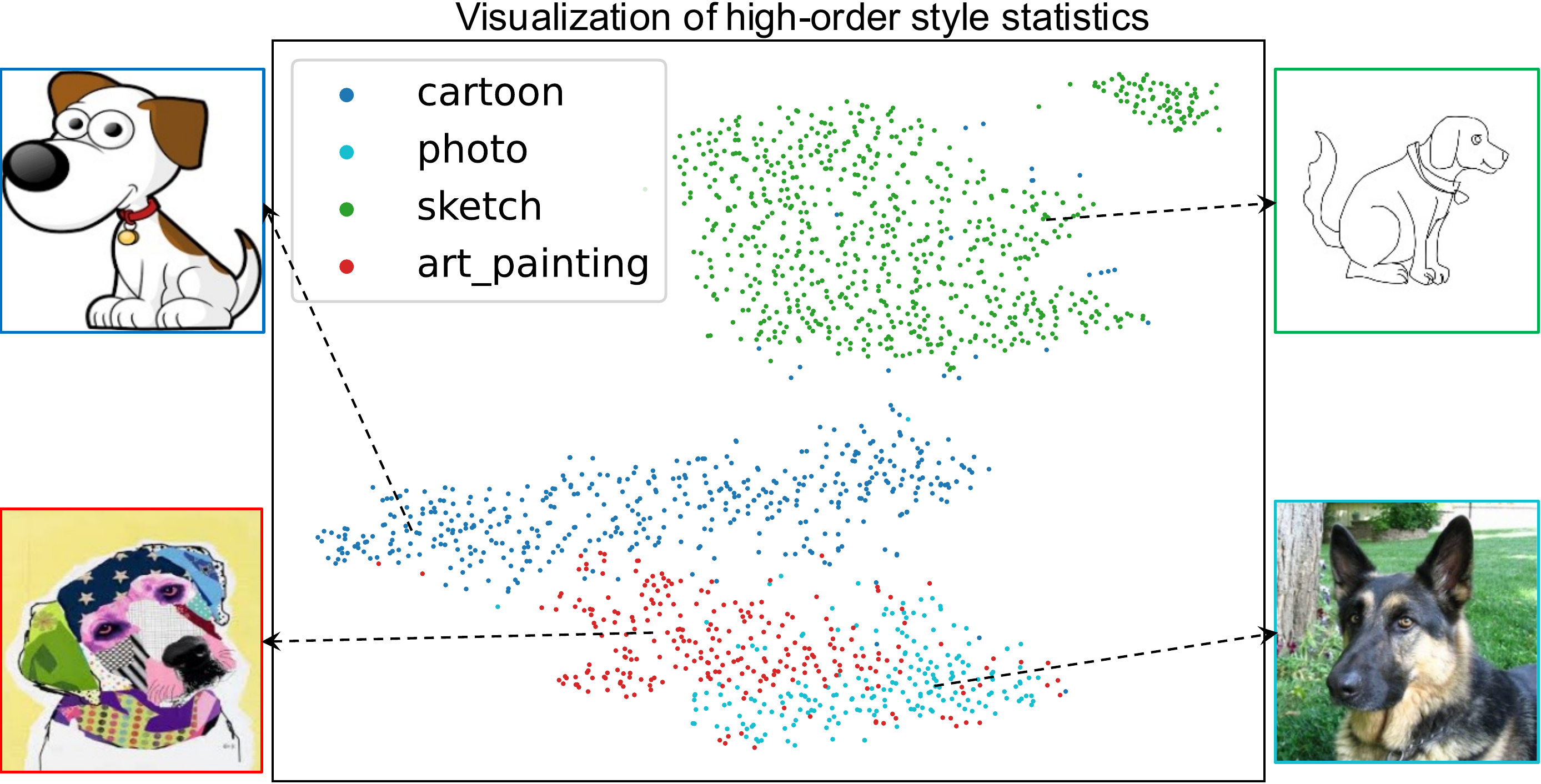}
		\caption{t-SNE \cite{van2008visualizing} visualization of the third standardized moment-skewness \cite{Skewness,joanes1998comparing}, which clearly shows that the style information can be represented by high-order statistics beyond mean and standard deviation. Besides skewness, style information can also be observed in the fourth standardized moment-kurtosis \cite{Kurtosis,joanes1998comparing} and infinite norm (please refer to the \textbf{supplementary file} for details).
			The visualized features are extracted from the 1st residual block of ResNet-18 \cite{he2016deep} trained on the dataset of four domains \cite{li2017deeper}. 
		} \label{Fig:vis_third_dog}
	\end{center}
	\vspace{-0.8cm}
    \end{figure}
	
	\noindent\textbf{EFDM for AST.}
	A simple encoder-decoder architecture is adopted, where we fix the encoder $f$ as the first few layers (up to $relu4\_1$) of a pre-trained VGG-19 \cite{simonyan2014very}. Given the content images $\mX$ and style images $\mY$, we first encode them to the feature space and apply EFDM to get the style-transferred features as:
	\begin{equation} \label{Equ:ast_EFDM}
	\mS = \textrm{EFDM}(f(\mX), f(\mY)).
	\end{equation} 
	Then, we train a randomly initialized decoder $g$ to map $\mS$ to the image space, resulting in the stylized images $g(\mS)$.
	Following \cite{huang2017arbitrary,dumoulin2016learned}, we train the decoder with the weighted combination of a content loss $\mathcal{L}_c$ and a style loss $\mathcal{L}_s$, leading to the following objective:
	\begin{equation} \label{Equ:ast_target}
	\mathcal{L} = \mathcal{L}_c  + \omega \mathcal{L}_s,
	\end{equation}
	where $\omega$ is a hyper-parameter balancing the two loss terms.
	Specifically, the content loss $\mathcal{L}_c$ is the Euclidean distance between  features of stylized images $f(g(\mS))$ and the style-transferred features $\mS$: 
	\begin{equation}
	\mathcal{L}_c = \| f(g(\mS)) -\mS \|_2.
	\end{equation}
	
	The style loss measures the distribution divergence between features of the stylized images $g(\mS)$ and style images $\mY$, which is instantiated as their divergence on mean and standard deviation in \cite{huang2017arbitrary} based on the Gaussian prior.
	%The style loss in \cite{huang2017arbitrary} only measures the divergence of mean and standard deviation based on the Gaussian prior. 
	To measure the distribution divergence more exactly, we introduce the style loss as the sum of Euclidean distance between features of the stylized images $\phi_i(g(\mS))$ and its style-transferred target $\textrm{EFDM}  (\phi_i(g(\mS)), \phi_i(\mY) )$: 
	\begin{equation} \label{Equ:style_loss}
	\mathcal{L}_s = \sum\nolimits_{i=1}^{L} \|\phi_i(g(\mS)) -   \textrm{EFDM}  (\phi_i(g(\mS)), \phi_i(\mY) )\|_2.
	\end{equation}
	Following \cite{huang2017arbitrary}, we instantiate $\{\phi_i\}_{i=1}^{L}$ as $relu1\_1$, $relu2\_1$, $relu3\_1$, and $relu4\_1$ layers in VGG-19.

	\noindent\textbf{EFDM for DG.} 
	Inspired by the studies that style information can be represented by the mean and standard deviation of image features \cite{huang2017arbitrary,li2017universal,lu2019closed}, Zhou \emph{et al.} \cite{zhou2021domain} proposed to generate style-transferred and content-preserved feature augmentations for DG problems. As we discussed before, distributions beyond Gaussian have high-order statistics other than mean and standard deviation, and hence the style information can be more accurately represented by using high-order feature statistics. The visualization in \cref{Fig:vis_third_dog} demonstrates that the third standardized moment-skewness \cite{Skewness,joanes1998comparing} can well represent the four different domains of the same object. This motivates us to utilize high-order statistics for feature augmentations in DG. 
	
	Since high-order feature statistics can be efficiently and implicitly matched via our proposed EFDM method, it is a natural idea to replace AdaIN with EFDM for cross-distribution feature augmentation in DG. To generate more diverse feature augmentations with mixed styles, following \cite{zhou2021domain} we extend the EFDM in \cref{Equ:sorting_matching_enable_gradient} by interpolating  sorted vectors, resulting in the Exact Feature Distribution Mixing (EFDMix) as:
	\begin{equation} \label{Equ:mixsorting}
	\textrm{EFDMix}(\vx,\vy):  \evo_{\tau_i} =  \evx_{\tau_i} + (1-\lambda) \evy_{\kappa_i}   - (1-\lambda)\langle \evx_{\tau_i} \rangle.
	\end{equation}
	The instance-wise mixing weight $\lambda$ is adopted and we sample $\lambda$ from the Beta-distribution: $\lambda \sim Beta(\alpha, \alpha)$, where $\alpha\in(0,\infty)$ is a hyper-parameter. We set $\alpha=0.1$ unless otherwise specified.  Obviously, EFDMix degenerates to EFDM when $\lambda=0$.
	
	Given the input feature $\mX$, following \cite{zhou2021domain} we adopt two strategies to mix with the style feature $\mY$. When domain labels are given, we sample $\mY$ from a domain different from that of $\mX$, leading to EFDMix w/ domain label. Otherwise, $\mY$ is obtained by shuffling $\mX$ along the batch dimension, resulting in EFDMix w/ random shuffle.
	We train the model solely with the cross-entropy loss. 
	Following \cite{zhou2021domain}, we insert the  EFDMix module to multiple lower-level layers, adopt a probability of $0.5$ to decide whether the EFDMix is activated in the forward pass of training stage, and deactivate it in the testing stage. 
	
	The advantage of utilizing high-order feature statistics could be intuitively clarified by the augmentation diversity. For example, given two different style features $\widehat{\vy}$ and $\widetilde{\vy}$ with the same mean and standard deviation and a specific mixing weight $\lambda$, the same augmented feature will be obtained by only utilizing the mean and standard deviation \cite{zhou2021domain}. 
	On the contrary, our EFDMix could generate two different augmentations by implicitly utilizing high-order statistics, resulting in more diverse feature augmentations.

	\begin{figure*}[htp]
		\begin{center}
			\includegraphics[width=0.99\textwidth]{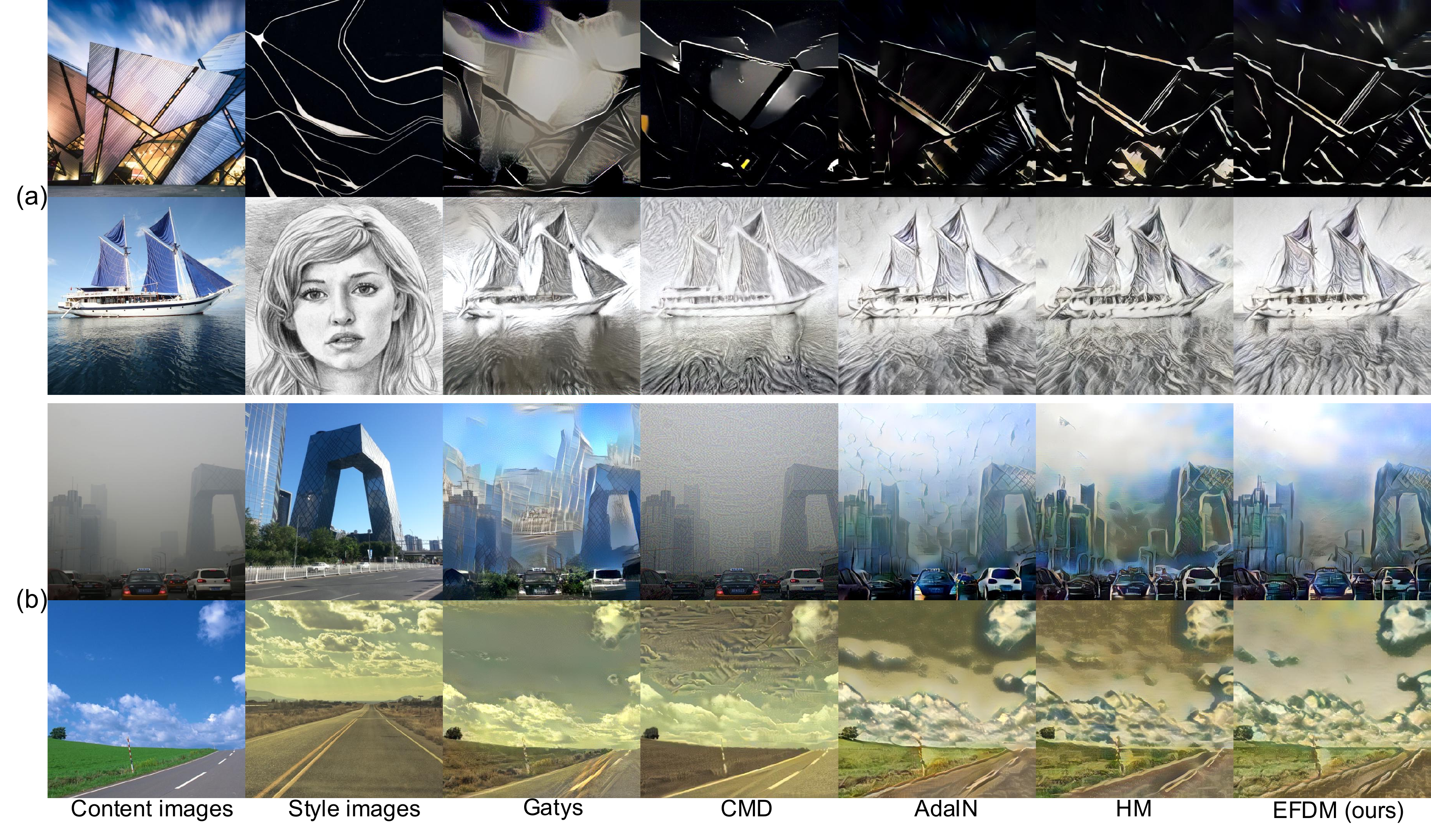}
		\end{center}
		\vspace{-0.5cm}
		\caption{Illustration of results on (a) style transfer \cite{huang2017arbitrary} (top two rows) and (b) the more challenging photo-realistic style transfer \cite{luan2017deep} (bottom two rows) tasks. Results of `Gatys' \cite{gatys2016image} and `CMD' \cite{kalischek2021light} are obtained with official codes. For HM, we use HM, instead of EHM, to approximately match eCDFs. More visualizations are provided in the \textbf{supplementary file}. 
		} \label{Fig:view_style}
		\vspace{-0.2cm}
	\end{figure*}
	\begin{figure*}[h]
		\begin{center}
			\includegraphics[width=0.98\linewidth]{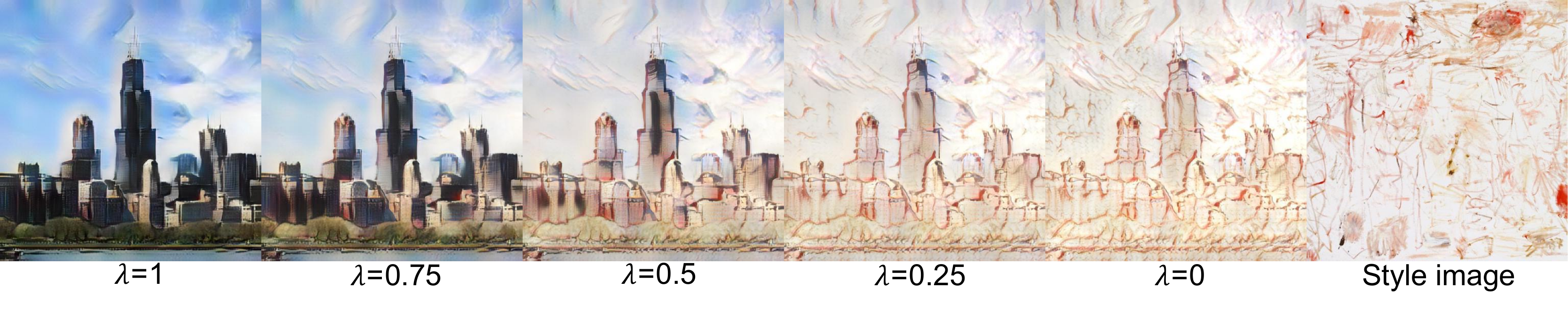}
		\end{center}
		\vspace{-0.5cm}
		\caption{Visualization of content-style trade-off with various $\lambda$ in \cref{Equ:mixsorting}. 
		} \label{Fig:more_vis_inter}
		\vspace{-0.5cm}
	\end{figure*}
	\begin{figure}[h]
		\begin{center}
			\includegraphics[width=0.95\linewidth]{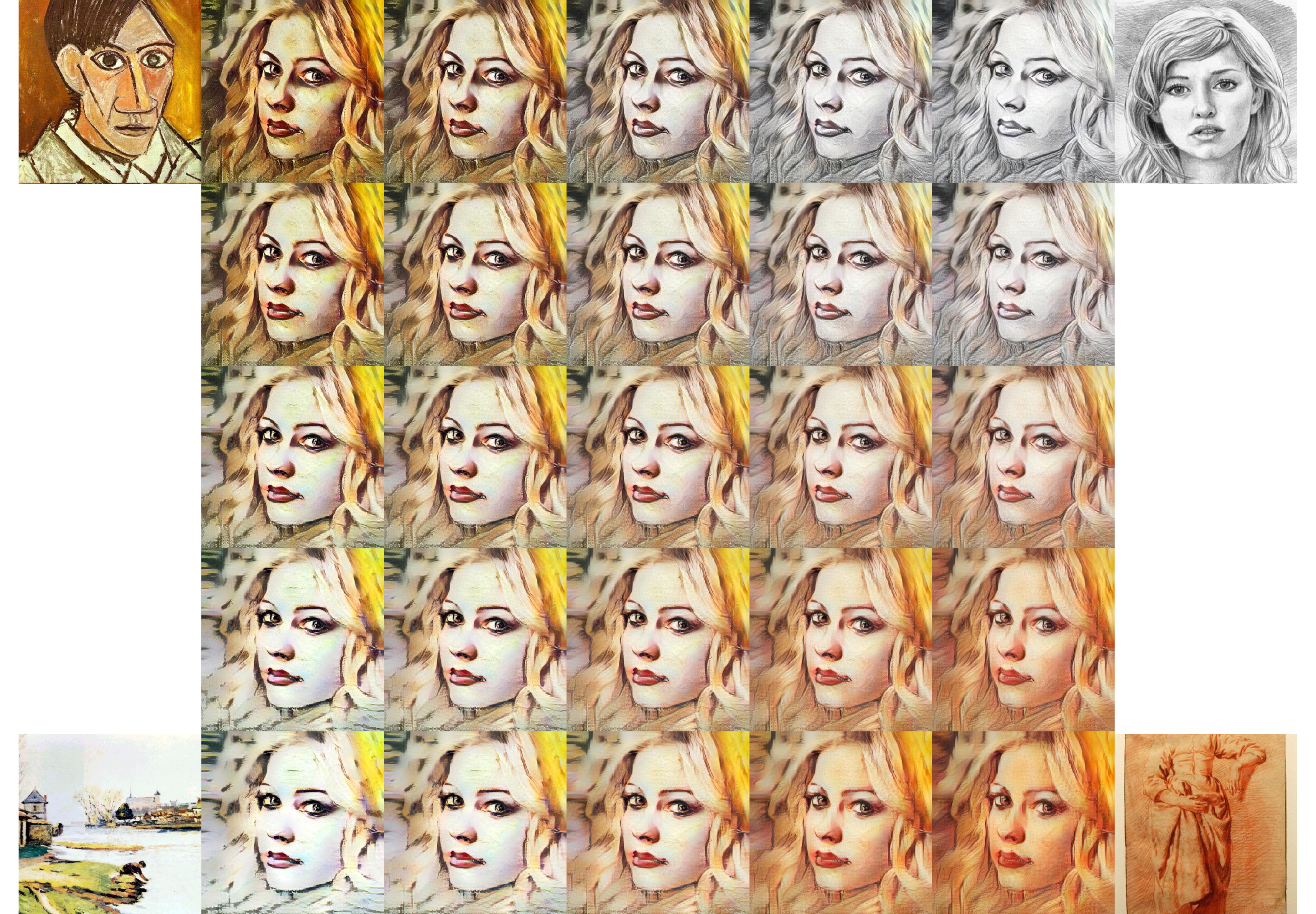}
		\end{center}
		\vspace{-0.5cm}
		\caption{Visualization of style interpolation.
		} \label{Fig:four_style}
		\vspace{-0.7cm}
	\end{figure}
	
	\vspace{-0.1cm}
	\section{Experiments} \label{Sec:experiments}
	\vspace{-0.1cm}
	We perform experiments on AST and DG tasks to validate the effectiveness of EFDM. 
	%The source codes of our algorithm are enclosed in the \textbf{supplementary file}. 
	\vspace{-0.1cm}
	\subsection{Experiments on AST}
	We closely follow \cite{huang2017arbitrary}  to conduct the experiments\footnote{\scriptsize{\url{https://github.com/naoto0804/pytorch-AdaIN}}} on AST. 
	Specifically, we adopt the adam optimizer, set the batch size as $8$ content-style image pairs, and set the hyper-parameter $\omega=10$.
	In training, the MS-COCO \cite{lin2014microsoft} and WikiArt \cite{wikiart} are adopted as the content and style images, respectively. We compare EFDM with state-of-the-arts in \cref{Fig:view_style}. One can see that our EFDM works stably across the style transfer (top two rows) and the more challenging photo-realistic style transfer (bottom two rows) tasks. By conducting feature distribution matching more exactly, it preserves more faithfully the image structures and details while transferring the style, and produces more photo-realistic results. In contrary, the competing methods may introduce many visual artifacts and image distortions. 
	%\textcolor{red}{Therefore, we should be vigilant for the application of EFDM in generating fake content.} 
	More visual results can be found in the \textbf{supplementary file}.
	
	\noindent\textbf{Content-style trade-off in the test stage.}
	The trade-off between content and style could be achieved by adjusting the hyper-parameter $\omega$ in \cref{Equ:ast_target}. Additionally, we could manipulate the content-style trade-off by interpolating between the content feature and style feature, which can be achieved with the EFDMix in \cref{Equ:mixsorting}.
	%With more smaller $\lambda$, the output image should be more closer to the target style. 
	The vanilla content image is expected when $\lambda=1$ and the model would output the most stylized image when $\lambda=0$. We illustrate an example in \cref{Fig:more_vis_inter}. We see that the images transition smoothly from the content style to target style by varying $\lambda$ from 1 to 0.

	\noindent\textbf{Style interpolation.}	Following \cite{huang2017arbitrary}, we interpolate feature maps to interpolate the $K$ style images $\mY_1$, $\mY_2$, $\cdots$, $\mY_K$ with corresponding weights $w_1,w_2,\cdots,w_K$ as follows:
	\begin{equation}
	g\left( \sum\nolimits_{k=1}^{K} w_k \textrm{EFDM}(\mX, \mY_k) \right),
	\end{equation}
	where $\sum_{k=1}^{K}w_k =1$. As illustrated in \cref{Fig:four_style}, new styles can be obtained by such style interpolation.

	\begin{table*}[t] \small
		\begin{center}
			\begin{tabular}{l| ccccc}
				\hline
				Method & Art & Cartoon & Photo & Sketch & Avg \\
				\hline
				\multicolumn{5}{c}{\textbf{Leave-one-domain-out generalization results}} \\
				\hline
				%MMD-AAE \cite{li2018domain} & 75.2 & 72.7 & 96.0 & 64.2 & 77.0 \\
				%CCSA   \cite{motiian2017unified}     & 80.5 & 76.9 & 93.6 & 66.8 & 79.4 \\
				JiGen  \cite{carlucci2019domain}      & 79.4 & 75.3 & 96.0 & 71.6 & 80.5 \\
				%CrossGrad \cite{shankar2018generalizing} & 79.8 & 76.8 & 96.0 & 70.2 & 80.7 \\
				%Epi-FCR  \cite{li2019episodic}  & 82.1 & 77.0 & 93.9 & 73.0 & 81.5 \\
				%Metareg  \cite{balaji2018metareg}  & 83.7 & 77.2 & 95.5 & 70.3 & 81.7 \\
				L2A-OT  \cite{zhou2020learning}   & 83.3 & 78.2 & 96.2 & 73.6 & 82.8 \\
				\hline
				ResNet-18  \cite{he2016deep}          & 77.0$\pm$0.6 & 75.9$\pm$0.6 & 96.0$\pm$0.1 & 69.2$\pm$0.6 & 79.5 \\
				%+ Manifold Mixup  \cite{verma2019manifold}  & 75.6$\pm$0.7 & 70.1$\pm$0.9 & 93.5$\pm$0.7 & 65.4$\pm$0.6 & 76.2 \\
				%+ Cutout  \cite{devries2017improved}             & 74.9$\pm$0.4 & 74.9$\pm$0.6 & 95.9$\pm$0.3 & 67.7$\pm$0.9 & 78.3 \\
				%+ CutMix \cite{yun2019cutmix}              & 74.6$\pm$0.7 & 71.8$\pm$0.6 & 95.6$\pm$0.4 & 65.3$\pm$0.8 & 76.8 \\
				%+ Mixup (w/o label interpolation) & 74.7$\pm$1.0 & 72.3$\pm$0.9 & 93.0$\pm$0.4 & 69.2$\pm$0.2 & 77.3 \\
				+ Mixup \cite{zhang2017mixup}               & 76.8$\pm$0.7 & 74.9$\pm$0.7 & 95.8$\pm$0.3 & 66.6$\pm$0.7 & 78.5 \\
				%+ DropBlock  \cite{ghiasi2018dropblock}        & 76.4$\pm$0.7 & 75.4$\pm$0.7 & 95.9$\pm$0.3 & 69.0$\pm$0.3 & 79.2 \\
				%+ MixStyle w/ random shuffle \cite{zhou2021domain}  & 82.3$\pm$0.2 & 79.0$\pm$0.3 & 96.3$\pm$0.3 & 73.8$\pm$0.9 & 82.8 \\
				%+ MixStyle w/ domain label \cite{zhou2021domain}   & 84.1$\pm$0.4 & 78.8$\pm$0.4 & 96.1$\pm$0.3 & 75.9$\pm$0.9 & 83.7 \\
				%+ MixStyle w/ random shuffle \cite{zhou2021domain}  & 82.4$\pm$0.2 & 79.4$\pm$0.8 & 96.2$\pm$0.1 & 72.3$\pm$0.6 & 82.6 \\
				+ MixStyle w/ domain label  \cite{zhou2021domain}  &83.1$\pm$0.8 & 78.6$\pm$0.9 & 95.9$\pm$0.4 & 74.2$\pm$2.7 & 82.9  \\
				%+ EFDMix w/ random shuffle (ours) & 83.2$\pm$0.7 & \textbf{79.8}$\pm$0.8 & 96.5$\pm$0.3 & 74.1$\pm$0.6 & 83.4 \\
				+ EFDMix w/ domain label  (ours)   & \textbf{83.9}$\pm$0.4 & \textbf{79.4}$\pm$0.7 & \textbf{96.8}$\pm$0.4 & \textbf{75.0}$\pm$0.7 & \textbf{83.9} \\
				\hline
				
				\hline
				ResNet-50 \cite{he2016deep}  & 84.4$\pm$0.9 & 77.1$\pm$1.4 & 97.6$\pm$0.2 & 70.8$\pm$0.7 & 82.5 \\
				%+ MixStyle w/ random shuffle \cite{zhou2021domain} & 88.7$\pm$0.7 & 81.4$\pm$0.7 & 98.0$\pm$0.2 & 75.0$\pm$0.6 & 85.8 \\
				+ MixStyle w/ domain label  \cite{zhou2021domain} & 90.3$\pm$0.3 & 82.3$\pm$0.7 & 97.7$\pm$0.4 & 74.7$\pm$0.7 & 86.2 \\
				%+ EFDMix  w/ random shuffle (ours) &  88.7$\pm$0.6 & 81.8$\pm$0.8 & 98.0$\pm$0.2 &  \textbf{77.7}$\pm$1.5 & 86.6 \\
				+ EFDMix  w/ domain label  (ours)  &   \textbf{90.6}$\pm$0.3 &  \textbf{82.5}$\pm$0.7 &  \textbf{98.1}$\pm$0.2 &  \textbf{76.4}$\pm$1.2 &  \textbf{86.9} \\
				% below is the results of applying to res1,2,3
				%+ MixStyle w/ random shuffle & 87.6$\pm$0.7 & 82.0$\pm$0.4 & 97.7$\pm$0.1 & 71.0$\pm$1.9 & 84.6 \\
				%+ MixStyle w/ domain label    & 88.1$\pm$0.6 & 82.6$\pm$0.9 & 97.5$\pm$0.3 & 72.7$\pm$1.1 & 85.1 \\
				%\hline
				%+ MixSort  w/ random shuffle & \textcolor{red}{87.1}$\pm$0.6 & 83.4$\pm$0.7 & \textbf{97.8}$\pm$0.1 & 76.7$\pm$1.5 & 86.3 \\
				%+ MixSort  w/ domain label    & \textbf{88.2}$\pm$1.0 & \textbf{83.5}$\pm$0.4 & \textbf{97.8}$\pm$0.2 & \textbf{77.6}$\pm$1.3 & \textbf{86.8} \\
				\hline
				\hline
				\multicolumn{5}{c}{\textbf{Single source generalization results}} \\
				\hline
				ResNet-18 \cite{he2016deep}  & 58.6$\pm$2.4 & 66.4$\pm$0.7 & 34.0$\pm$1.8 & 27.5$\pm$4.3 & 46.6  \\
				+ MixStyle w/ random shuffle \cite{zhou2021domain} & 61.9$\pm$2.2 & 71.5$\pm$0.8 & 41.2$\pm$1.8 & 32.2$\pm$4.1 & 51.7   \\
				+ EFDMix w/ random shuffle (ours) &  \textbf{63.2}$\pm$2.3 & \textbf{73.9}$\pm$0.7 & \textbf{42.5}$\pm$1.8 & \textbf{38.1}$\pm$3.7 & \textbf{54.4} \\
				\hline
				
				\hline
				ResNet-50 \cite{he2016deep}  & 63.5$\pm$1.3 & 69.2$\pm$1.6    & 38.0$\pm$0.9 & 31.4$\pm$1.5 & 50.5 \\
				%% +res12
				%+ MixStyle w/ random shuffle & 67.5$\pm$1.5 & 75.2$\pm$1.3 & 42.8$\pm$0.8 & 36.4$\pm$1.2 & 55.5 \\
				%+ EFDMix  w/ random shuffle &  \textbf{68.5}$\pm$1.1 &  \textbf{76.4}$\pm$0.9 &  \textbf{43.5}$\pm$0.5 &  \textbf{39.2}$\pm$1.3 &  \textbf{56.9} \\
				%%  +res123
				+ MixStyle w/ random shuffle  \cite{zhou2021domain} & 73.2$\pm$1.1 & 74.8$\pm$1.1 & 46.0$\pm$2.0 & 40.6$\pm$2.0 &  58.6 \\
				+ EFDMix  w/ random shuffle (ours) & \textbf{75.3}$\pm$0.9 & \textbf{77.4}$\pm$0.8 & \textbf{48.0}$\pm$0.9 & \textbf{44.2}$\pm$2.4 &  \textbf{61.2}  \\
				\hline
			\end{tabular}
			\vspace{-0.1cm}
			\caption{Domain generalization results of category classification on PACS. Results of MixStyle are obtained with official codes. The listed domain is the test domain in the leave-one-domain-out setting, while it is the training one in the single source generalization setting. } 
			\label{tab:321_pacs}
		\end{center}
		\vspace{-0.5cm}
	\end{table*}

	\begin{table*}[h] \small
	\begin{center}
		\begin{tabular}{L{45.6mm}|C{10.0mm}C{10.0mm}C{10.0mm}C{10.0mm}|C{10.0mm}C{10.0mm}C{10.0mm}C{10.0mm}}
			\hline
			\multirow{2}{*}{Methods} & \multicolumn{4}{c|}{MarKet1501$\to$GRID} & \multicolumn{4}{c}{GRID$\to$MarKet1501} \\
			& mAP & R1 & R5 & R10 & mAP & R1 & R5 & R10 \\
			\hline
			OSNet  \cite{zhou2019omni} & 33.3$\pm$0.4 & 24.5$\pm$0.4 & 42.1$\pm$1.0 & 48.8$\pm$0.7 & 3.9$\pm$0.4 & 13.1$\pm$1.0 & 25.3$\pm$2.2 & 31.7$\pm$2.0 \\ 
			+ MixStyle w/ random shuffle  \cite{zhou2021domain} &  33.8$\pm$0.9 & 24.8$\pm$1.6 & 43.7$\pm$2.0 & 53.1$\pm$1.6 & 4.9$\pm$0.2 & 15.4$\pm$1.2 & 28.4$\pm$1.3 & 35.7$\pm$0.9  \\
			+ EFDMix w/ random shuffle (ours) & \textbf{35.5}$\pm$1.8 & \textbf{26.7}$\pm$3.3 & \textbf{44.4}$\pm$0.8 & \textbf{53.6}$\pm$2.0 &  \textbf{6.4}$\pm$0.2 & \textbf{19.9}$\pm$0.6 & \textbf{34.4}$\pm$1.0 & \textbf{42.2}$\pm$0.8  \\
			\hline
		\end{tabular}
		\vspace{-0.1cm}
		\caption{Domain generalization results on the cross-domain person re-ID task.  Results of MixStyle are obtained with official codes.} % res12
		\label{Tab:reid}
	\end{center}
	\vspace{-0.7cm}
\end{table*}

	\subsection{Experiments on DG}	We closely follow MixStyle \cite{zhou2021domain} to conduct experiments on DG\footnote{\scriptsize{\url{https://github.com/KaiyangZhou/mixstyle-release}}}, including data preparing, model training and selection. In other words, we only replace the MixStyle module with EFDMix, which is detailed as follows.
	
	\noindent\textbf{Generalization on category classification.} 
	We adopt the popular DG benchmark dataset of PACS \cite{li2017deeper}, which includes $9,991$ images shared by $7$ classes and $4$ domains, \ie, Art, Cartoon, Photo, and Sketch. Two task settings are adopted. In the leave-one-domain-out setting \cite{li2017deeper}, we train the model on three domains and test on the remaining one.  In the single source DG \cite{volpi2018generalizing, qiao2020learning}, models are trained on one domain and tested on the remaining three. 
	We adopt ResNet-18 and ResNet-50, which are pre-trained on the ImageNet dataset, as the backbones.

	We compare our method with the latest state-of-the-art MixStyle \cite{zhou2021domain}, the regularization based methods \cite{zhang2017mixup,verma2019manifold,ghiasi2018dropblock,yun2019cutmix,devries2017improved} and the representative DG methods \cite{motiian2017unified,li2018domain,carlucci2019domain,shankar2018generalizing,balaji2018metareg,li2019episodic,zhou2020learning}. Due to the limit of space, only partial results are reported in \cref{tab:321_pacs} and more comprehensive results are given in the \textbf{supplementary file}.
	One can see that our EFDMix consistently outperforms MixStyle, as well as other competing methods, on both settings. More advantages over the competing methods can be observed on the single source generalization setting, where the training data have less diversity. This can be explained by the more diverse  feature augmentations via EFDMix, as clarified in \cref{subsec:apply_sort_matching}. 
	
	We note there are different experimental strategies in the DG community. Following the recent DomainBed \cite{gulrajani2020search}, our EFDMix achieves 87.9\% accuracy on the PACS dataset, outperforming the strong ERM benchmark \cite{gulrajani2020search} by 1.2\%. Please refer to the \textbf{supplementary file} for details.

	\begin{figure*}[h!]
		\centering
		\includegraphics[width=0.95\textwidth]{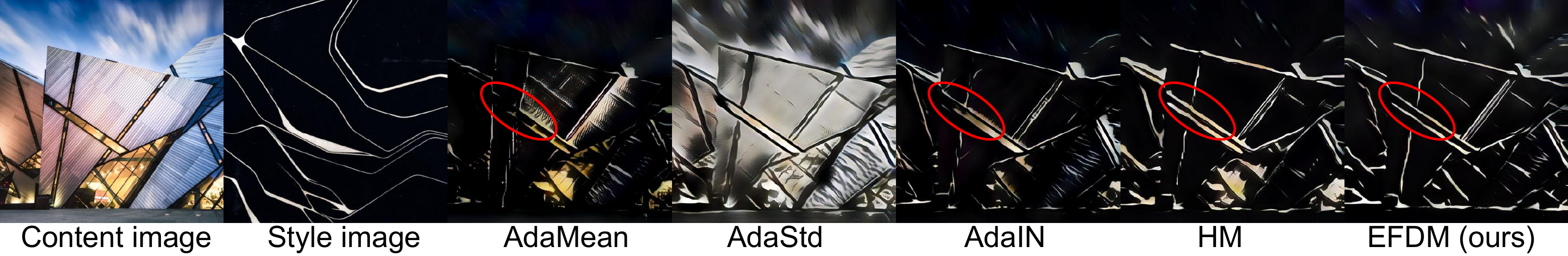}
		\vspace{-0.3cm}
		\caption{Qualitative analyses on the role of different orders of feature statistics on AST.} \label{Fig:qualitative}
		\vspace{-0.5cm}
	\end{figure*}

	\noindent\textbf{Generalization on instance retrieval.} We adopt the person re-identification (re-ID) datasets of Markert1501 \cite{zheng2015scalable} and GRID \cite{loy2009multi}
	%Duke \cite{ristani2016performance,zheng2017unlabeled} 
	to conduct cross-domain instance retrieval. We follow \cite{zhou2021domain} to conduct experiments with the OSNet \cite{zhou2019omni}. Similar to the findings in classification, EFDMix outperforms MixStyle and other competitors, as shown in  \cref{Tab:reid}. This once again validates the effectiveness of utilizing high-order statistics for feature augmentations in DG.
	%better to add some discussions here. 
	%To evaluate the model generalization across distributions, we train the model on one dataset and test it on the other, where each camera view is a distinct domain. 
	%and measure performance in terms of mean average precision (mAP) and ranking accuracy. 
	%We insert MixSort after the $1$st and $2$nd residual blocks. 
	
	\vspace{-0.1cm}
	\subsection{Discussions} \label{subsec:analysis}
	\vspace{-0.1cm}
	
\begin{figure}
		\centering
		\includegraphics[width=0.6\linewidth]{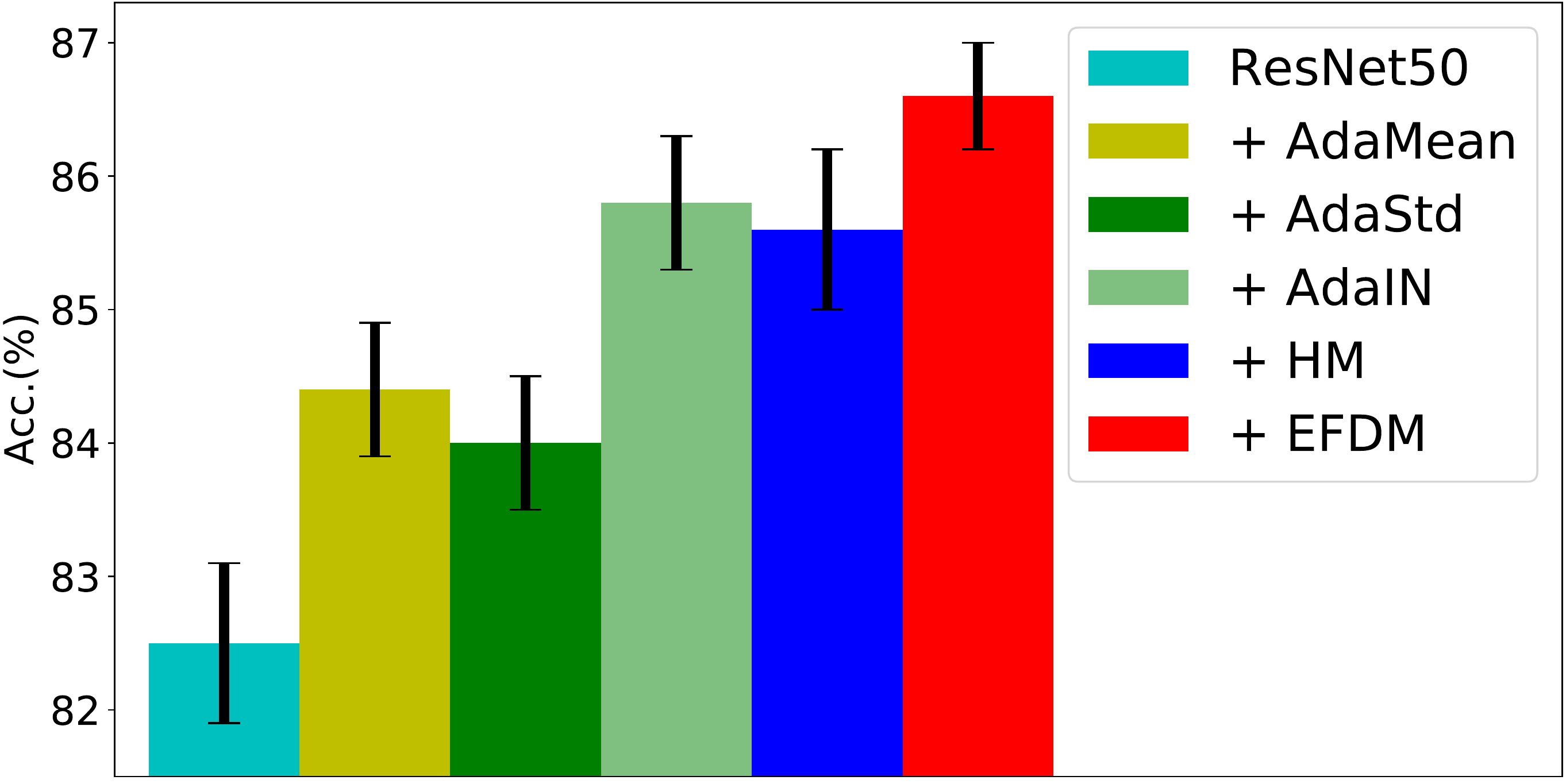}
		\vspace{-0.2cm}
		\caption{Quantitative analyses on the role of different orders of feature statistics on PACS dataset.} \label{Fig:quantitative}
		\vspace{-0.3cm}
	\end{figure}

	\noindent \textbf{The role of different orders of feature statistics. } To make further investigation on the role of different orders of feature statistics, we implement AdaIN by matching only feature mean and standard deviation, resulting in the AdaMean and AdaStd variants of it. (Please refer to the \textbf{supplementary file} for details.) The qualitative results on AST and the quantitative results on DG are illustrated in \cref{Fig:qualitative} and \cref{Fig:quantitative}, respectively. From \cref{Fig:qualitative}, we can see that AdaMean roughly matches the basic color tone. AdaStd preserves the structure of the content image but with wrong color tone. By matching both mean and standard deviation, AdaIN preserves more details and correct tone. 
	With the implicitly matched high-order feature statistics, EFDM preserves the most content details. 
	From \cref{Fig:quantitative}, we see that performing feature augmentation with either AdaMean or AdaStd could improve over the ResNet-50 baseline, while AdaMean performs slightly better. AdaIN outperforms AdaMean and AdaStd by more than $1\%$, justifying the effectiveness of utilizing more feature statistics. By matching high-order feature statistics implicitly, EFDM achieves the best result. Though HM approximately matches eCDFs, it cannot even ensure the exact matching of mean and standard deviation, leading to degenerated performance. 

	\noindent 	\textbf{User study on style transfer.} As shown in \cref{Tab:speed}, our method receives the most votes for its better stylized performance across competing AST methods. Please refer to the \textbf{supplementary file} for more details.
	
\begin{figure}
		\centering
		\includegraphics[width=0.6\linewidth]{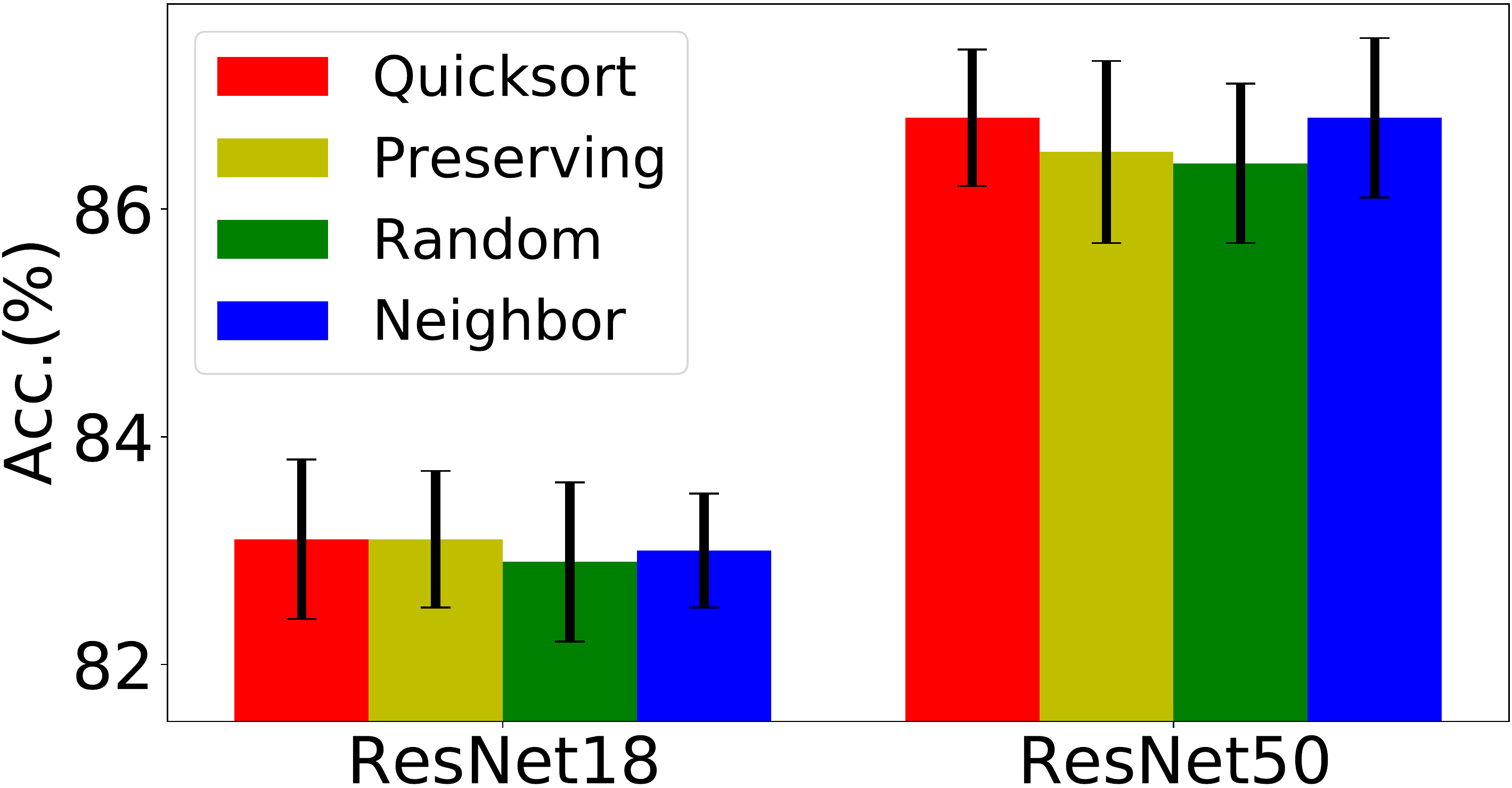}
		\vspace{-0.25cm}
		\caption{Results on PACS dataset with different implementations of EFDM. Besides Sort-Matching with Quicksort \cite{sedgewick1978implementing}, we preserve the order of equivalent values in $\vx$ (Preserving), randomly sort equivalent values \cite{rosenfeld1976digital} (Random), and sort equivalent values according to their local mean \cite{hall1974almost} (Neighbor) to implement EFDM, respectively. We see that the different implementations lead to similar results.} \label{Fig:order_strategies}
		\vspace{-0.4cm}
	\end{figure}

	\noindent \textbf{EFDM with different EHM algorithms.}
	Different EHM algorithms are distinguished by their sort strategies of equivalent values. In \cref{Fig:order_strategies}, we implement EFDM with different EHM algorithms on the task of DG. One can see that they yield similar accuracies on the PACS dataset. Considering that the quicksort-empowered Sort-Matching has the fastest speed, we adopt it for EFDM in our work.

	\begin{table} \small
		\centering
		\begin{tabular}{c|ccccc}
			\hline
			Methods  & Gatys \cite{gatys2016image} & CMD \cite{kalischek2021light} & HM & EFDM & AdaIN  \\
			\hline
			Time (s)  & 25.61 & 19.84 & 0.33 & 0.0039 & 0.0038 \\
			Percent (\%) & 19.9 & 12.2 & 17.1 & 29.8 & 21.0  \\
			\hline
		\end{tabular}
		\vspace{-0.3cm}
		\captionof{table}{Average running time and user preference across competing AST methods. The running time is averaged 
			on  512$\times$512 images with a Tesla V100 GPU.  
		} \label{Tab:speed}
		\vspace{-0.6cm}
	\end{table}

	\noindent\textbf{Running time.} We evaluate the speed of our EFDM method on the AST task. The average running time of different algorithms to process a 512$\times$512 image is listed in \cref{Tab:speed}. EFDM is significantly faster than methods in \cite{gatys2016image,kalischek2021light} and the HM based algorithm, which is implemented with the skimage library\footnote{\url{https://github.com/scikit-image/scikit-image}}.
	It has nearly the same speed as the seminal AdaIN \cite{huang2017arbitrary} and runs at 256 FPS for images of size 512 $\times$ 512, making it  applicable for real-time applications. 
	
	\noindent\textbf{Limitations.} Compared to AdaIN \cite{huang2017arbitrary} with linear complexity, EFDM has a higher complexity of $n\log(n)$. Fortunately, due to the finite feature size, its running time is comparable to AdaIN on the AST and DG tasks. In addition, following \cite{huang2017arbitrary,kalischek2021light,zhou2021domain}, we assume that different feature channels are independent, which is not exactly true and is challenged by \cite{li2017universal,lu2019closed}.

	\noindent More discussions on EFDMix, EFDM vs. EFDMix, the selection of $\alpha$ in \cref{Equ:mixsorting}, loss curves, the influence of ReLU functions, the comparison to related methods on DG \cite{zhou2021domain,tang2021crossnorm,jin2020style,fan2021adversarially}, and the detailed analysis on computation time can be found in the \textbf{supplementary file}.
	
	\vspace{-0.2cm}
	\section{Conclusion}
	\vspace{-0.1cm}
	We made the first attempt, to our best knowledge, to perform exact matching of feature distributions, and applied the so-called exact feature distribution matching (EFDM) method to applications of AST and DG. We employed a fast EHM algorithm, \ie, Sort-Matching, to implement EFDM in the deep feature space. The proposed EFDM method demonstrated superior performance to existing state-of-the-arts of AST and DG in terms of visual quality and quantitative measures. Our work opened a door to perform EFDM for visual learning tasks efficiently. Extensive investigations could be followed up, \eg, empowering classical normalization \cite{ioffe2015batch} beyond statistics of mean and standard deviation.  
	%\noindent\small\textbf{Acknowledgment. \textcolor{red}{XXX}}

	%%%%%%%%% REFERENCES
	{\small
		\bibliographystyle{ieee_fullname}
		\bibliography{egbib}
	}
	
\end{document}